\documentclass[letterpaper]{article} 
\PassOptionsToPackage{table}{xcolor}
\usepackage{xcolor}
\usepackage{aaai2026}  
\usepackage{times}  
\usepackage{helvet}  
\usepackage{courier}  
\usepackage[hyphens]{url}  
\usepackage{graphicx} 
\usepackage{natbib}  
\usepackage{caption} 
\usepackage{adjustbox}
\usepackage{multirow}
\usepackage{amsfonts}
\usepackage{booktabs}
\usepackage{amsmath}
\usepackage{cuted}
\usepackage{pgfplots}
\usepackage{tabularx}

\usepackage{algorithm}
\usepackage{algorithmic}
\usepackage[most]{tcolorbox}
\usepackage[T1]{fontenc}
\usepackage{textcomp} 
\usepackage{newfloat}
\usepackage{listings}
\DeclareCaptionStyle{ruled}{labelfont=normalfont,labelsep=colon,strut=off} 
\floatstyle{ruled}
\newfloat{listing}{tb}{lst}{}
\floatname{listing}{Listing}
\title{
\texttt{SageLM}\includegraphics[height=1em]{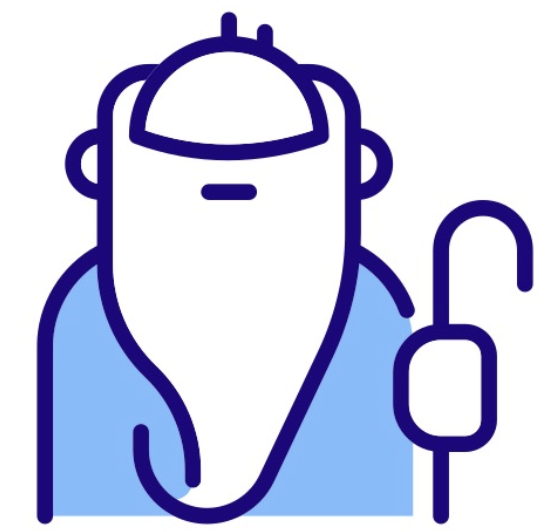}: A Multi-aspect and Explainable Large Language Model \\for Speech Judgement
}

\author{
    Yuan Ge\textsuperscript{1, $\dagger$}, Junxiang Zhang\textsuperscript{1, $\dagger$}, Xiaoqian Liu\textsuperscript{1}, Bei Li\textsuperscript{2}, Xiangnan Ma\textsuperscript{1}, \\Chenglong Wang\textsuperscript{1}, Kaiyang Ye\textsuperscript{1}, Yangfan Du\textsuperscript{1}, Linfeng Zhang\textsuperscript{3, $\ddagger$}, \\Yuxin Huang\textsuperscript{4}, Tong Xiao\textsuperscript{1, 5, $\ddagger$}, Zhengtao Yu\textsuperscript{4}, Jingbo Zhu\textsuperscript{1, 5}
}

\affiliations{
    \textsuperscript{\rm 1}Northeastern University, China\\
    \textsuperscript{\rm 2}Meituan\\
    \textsuperscript{\rm 3}Shanghai Jiao Tong University\\
    \textsuperscript{\rm 4}Kunming University of Science and Technology\\
    \textsuperscript{\rm 5}NiuTrans Research
}

\usepackage{bibentry}

\begin{document}

\maketitle

\begin{abstract}

Speech-to-Speech (S2S) Large Language Models (LLMs) are foundational to natural human-computer interaction, enabling end-to-end spoken dialogue systems. 
However, evaluating these models remains a fundamental challenge. 
We propose \texttt{SageLM}, an end-to-end, multi-aspect, and explainable speech LLM for comprehensive S2S LLMs evaluation. 
First, unlike cascaded approaches that disregard acoustic features, SageLM jointly assesses both semantic and acoustic dimensions. 
Second, it leverages rationale-based supervision to enhance explainability and guide model learning, achieving superior alignment with evaluation outcomes compared to rule-based reinforcement learning methods. 
Third, we introduce \textit{SpeechFeedback}, a synthetic preference dataset, and employ a two-stage training paradigm to mitigate the scarcity of speech preference data. 
Trained on both semantic and acoustic dimensions, SageLM achieves an 82.79\% agreement rate with human evaluators, outperforming cascaded and SLM-based baselines by at least 7.42\% and 26.20\%, respectively.
{
    \renewcommand{\thefootnote}{}%
    \footnotetext{$\dagger$ Equal contribution.}
    \footnotetext{$\ddagger$ Corresponding author.}
    \footnotetext{~~~\url{https://github.com/IronBeliever/SageLM}}
}

\end{abstract}

\section{Introduction}

The advent of large language models (LLMs) has revolutionized human-computer interaction, yet the ultimate goal remains a seamless, natural dialogue that mirrors human conversation \cite{clark1996using,luger2016like}. The next frontier in this pursuit is speech-to-speech (S2S) interaction, where the nuances of communication extend far beyond mere semantic content \cite{levitan2015entrainment,porcheron2018voice}. In human dialogue, how something is said is often as important as what is said \cite{pierrehumbert1990meaning,hirschberg2002pragmatics}. For instance, the query, ``You know what? I won a million dollar lottery today!'' demands more than a semantically correct response; it calls for an expression of incredulous surprise or shared joy. This inextricable coupling of semantics and acoustics makes evaluating S2S dialogue systems a profound and unsolved challenge.

\begin{figure}[t]
\centering
\includegraphics[width=0.47\textwidth]{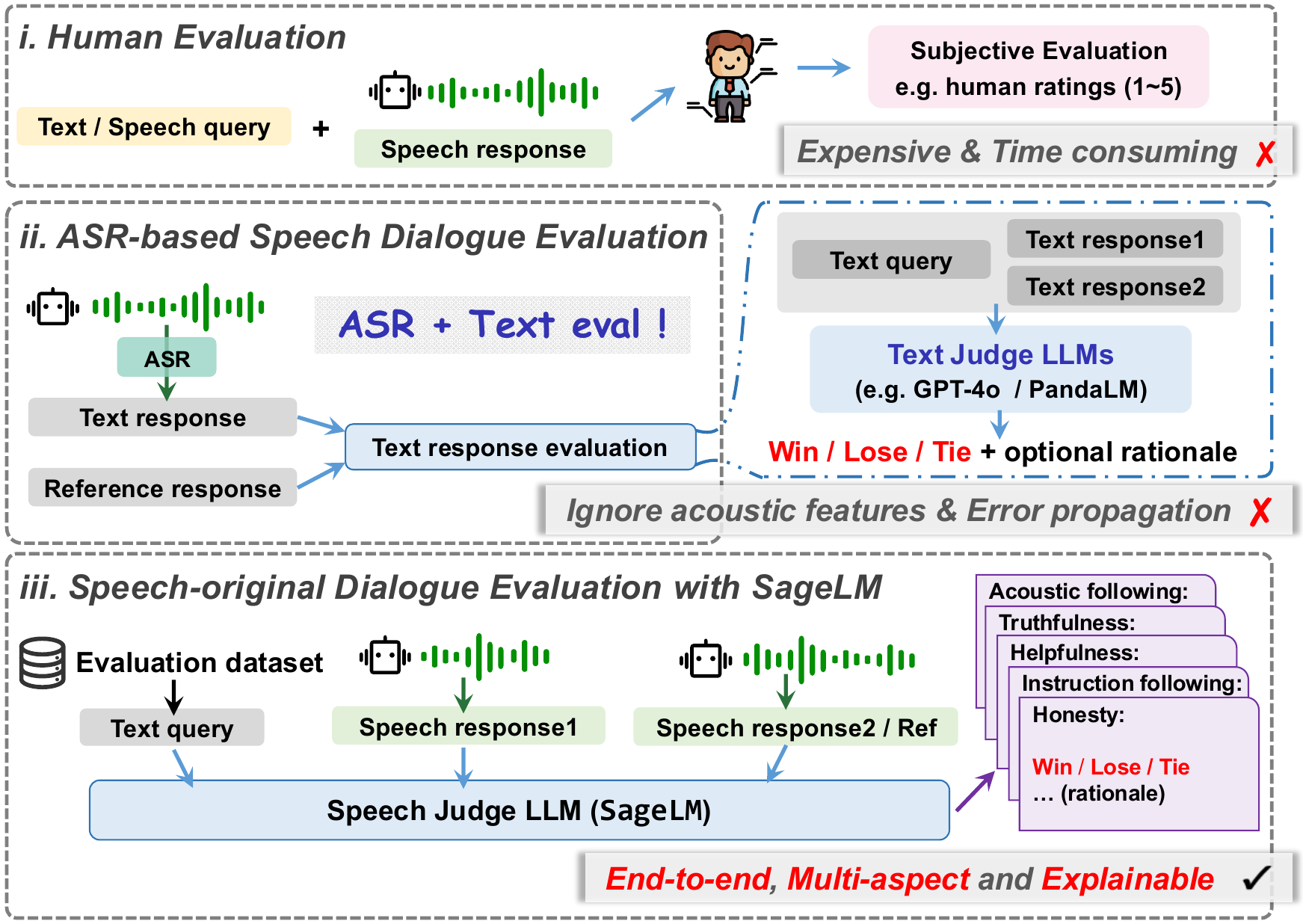} 
\caption{Recent speech-to-speech LLMs evaluation methods rely on human annotations or cascaded pipelines. We propose \texttt{SageLM}, an end-to-end speech dialogue evaluator that provides explainable judgment results across five aspects, including both semantic and acoustic dimensions.
}
\label{fig:motivation}
\end{figure}

Current evaluation paradigms are fundamentally inadequate for this task. The predominant approach, \textit{a cascaded pipeline} that first transcribes speech with an Automatic Speech Recognition (ASR) model and then uses a text-based LLM for judgment, is critically flawed \citep{nachmani2024spoken, defossez2024moshi, zhang2024omniflatten, chen2024slam}. Not only is it susceptible to cascading ASR errors, but it is entirely blind to the acoustic dimension, failing to assess the appropriateness of tone, emotion, or prosody. The alternative, \textit{human evaluation}, serves as a gold standard but is prohibitively expensive and slow, rendering it impractical for the rapid, large-scale iteration required to develop advanced S2S models \citep{veluri-etal-2024-beyond, ding2025kimi}. This leaves a critical gap in the field: the absence of a scalable, automated evaluator that can holistically judge both the content and character of spoken dialogue.

To fill this void, we introduce \texttt{SageLM}, a multi-a\textbf{s}pect and expl\textbf{a}inable jud\textbf{ge} \textbf{l}anguage \textbf{m}odel designed specifically for the nuanced demands of S2S dialogue evaluation. 
Building on the ``LLM-as-a-judge'' paradigm \cite{zheng2024judging, li2024generation, gu2024survey}, SageLM operates end-to-end, directly processing speech to render a holistic judgment, thereby bypassing the fragile ASR pipeline. Building such a sophisticated judge, however, presents two foundational obstacles: a data bottleneck and a methodological dilemma.

First, the lack of a large-scale, annotated speech preference dataset has been a primary barrier to progress. To overcome this, we construct \textit{SpeechFeedback}, a comprehensive and diverse preference dataset of 324,774 instances. Each instance contains a speech query, a pair of contrasting responses, and detailed preference annotations covering both semantic relevance and acoustic quality, providing the necessary foundation for training a nuanced speech judge.

Second, even with the right data, the training methodology is paramount.
Recent trends have explored rule-based Reinforcement Learning (RL) for tasks where solutions are easily verified \cite{guo2025deepseek, swamy2025all, li2025reinforcement}. However, we find this approach fundamentally misaligned with our goal. A simple, rule-based reward signal is insufficient for the complex reasoning required in dialogue evaluation, making the model prone to reward hacking and, crucially, failing to enforce \textit{consistency between its judgments and the explanatory rationales}.
We demonstrate that a supervised fine-tuning approach on LLM-annotated rationales is superior enough. This method compels the model not only to predict a judgment but to reason its way to that conclusion, fostering a deeper and more robust understanding that aligns the ``what'' with the ``why.''
Experimental results demonstrate the success of this approach. SageLM achieves an 82.79\% agreement rate with human judgments, significantly outperforming strong cascaded baselines like Whisper + GPT-4o by 7.42\% and other end-to-end speech models by a remarkable 26.20\%. 
Our contributions are:
\begin{itemize}
    \item  We construct \textit{SpeechFeedback}, the first large-scale, multi-aspect speech preference dataset to facilitate research in S2S evaluation.
    \item We demonstrate that for complex judgment tasks, fine-tuning on explicit rationales is a more effective method than rule-based RL, leading to better performance and reasoning consistency.
    \item We introduce \texttt{SageLM}, an end-to-end, explainable judge for S2S dialogue that significantly surpasses existing cascaded and integrated evaluation models, establishing a new state-of-the-art.
    \item We validate SageLM's effectiveness in rigorous experiments, showing high agreement with human evaluators and superior performance against formidable baselines.
\end{itemize}

\section{Related Work}

\subsection{Speech Large Language Models} 
Speech Large Language Models (SLMs) are typically categorized into speech-to-speech (S2S) and speech-to-text (S2T) LLMs.
Since GPT-4o, end-to-end S2S LLMs have gained increased attention  \cite{defossez2024moshi,zhang2024omniflatten,chen2024slam,ding2025kimi,goel2025audio}.
Despite rapid advancements in S2S LLMs, accurately evaluating their dialogue ability remains a significant challenge.
S2T LLMs combine the language modeling capabilities of text-based LLMs with speech understanding capabilities of audio encoders \cite{chu2024qwen2audio, zhang2025soundwave} or codecs \cite{zhang2023speechgpt, zhang2024omniflatten, zhan2024anygpt, li2025baichuan}. 
This integration enables S2T LLMs to perform speech dialogue understaning tasks.

\subsection{Text Large Language Models Evaluation} 
The evaluation of text-based LLMs in conversational settings primarily assesses their instruction following capabilities, employing either human or model-based evaluation paradigms.
The most straightforward approach involves human annotators assigning point-wise or pair-wise subjective labels. However, human evaluation is costly and time-consuming. 
To address this, the \textit{LLMs-as-a-judge} paradigm leverages powerful models such as GPT-4 to evaluate candidate responses \citep{zhou2023lima, rafailov2023direct, dubois2023alpacafarm, lee2023rlaif}. Despite its efficiency, this approach introduces concerns such as model bias \citep{zheng2024judging, wang-etal-2024-large-language-models-fair}, privacy risks, and computational cost. Consequently, recent open-source efforts have focused on instruction-tuning pretrained LLMs to enhance their evaluation capabilities \citep{pandalm2024,li2023generative}, aiming to mitigate bias, reduce costs, and preserve user privacy. Other works explore lightweight evaluation methods, utilizing smaller models to assess response quality more efficiently \citep{ge-etal-2024-clustering,sinha-etal-2020-learning,phy-etal-2020-deconstruct}.

\begin{figure*}[t]
\centering
\includegraphics[width=0.95\textwidth]{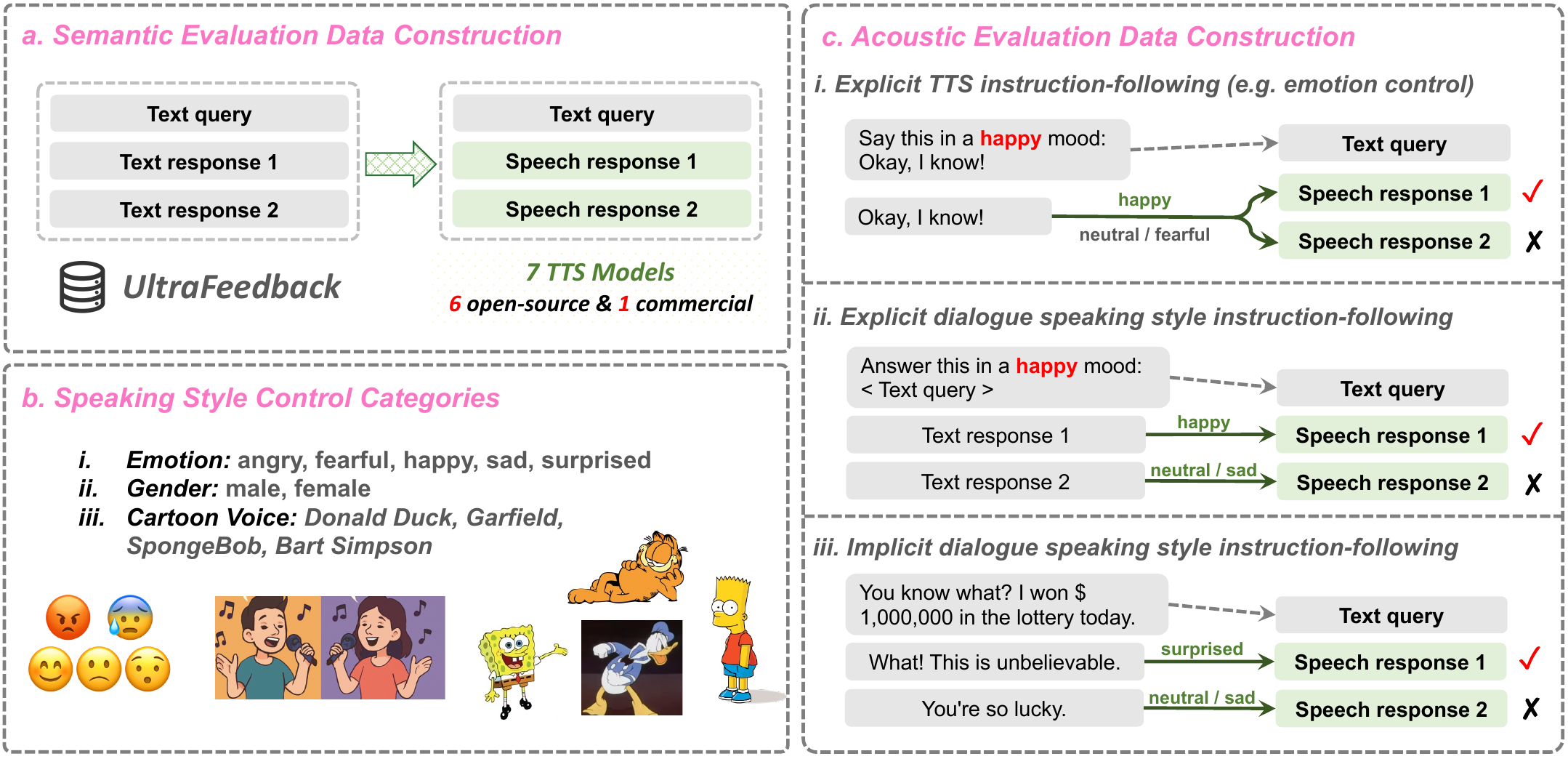} 
\caption{Data construction pipeline of \textit{SpeechFeedback}.}
\label{fig:data construction}
\end{figure*}

\subsection{S2S Large Language Models Evaluation} 
Evaluating S2S LLMs remains an open challenge. 
The modalities supported by S2S LLMs are determined by their architectural design. Typically, S2S models, such as Kimi-Audio \cite{ding2025kimi}, support both text and speech as inputs and outputs. 
Evaluation in this comprehensive setting encompasses fundamental speech capabilities, audio understanding, and speech conversation.
The evaluation of fundamental speech capabilities includes ASR and TTS tasks, commonly measured using established metrics like Word Error Rate (WER) and Mean Opinion Score (MOS). 
Audio understanding tasks focus on the model’s ability to comprehend and reason about both semantic and acoustic information within audio, utilizing benchmarks such as MMAU \cite{sakshi2024mmau}, ClothoAQA \cite{lipping2022clotho}, and AIR-Bench \cite{yang2024air}.
However, speech conversation evaluation emphasizes the model’s dialog capabilities, including audio-to-text chat and speech-to-speech chat. This demanding task requires not only an understanding of audio semantics and paralinguistic cues, but also competencies in language modeling, reasoning, and speaking style control. Notably, evaluating speech-to-speech conversation remains an open challenge.
As shown in Fig. \ref{fig:motivation}, human evaluation, while considered the gold standard, is costly and time-consuming, making it impractical to scale comprehensive benchmarking. Furthermore, human evaluation typically overlooks explainability due to the additional cost and time required.
ASR-based evaluation, which transcribes the model's speech responses into text followed by conventional text-based LLM conversation evaluation, introduces compounding errors because even state-of-the-art ASR models exhibit non-negligible WERs. Moreover, such cascaded methods disregard paralinguistic features (e.g., emotion, speech rate), thereby failing to evaluate the model's ability to adhere to style-related instructions in speech generation.

\section{Speech Dialogue Judgment}
This section defines the speech dialogue judgment task and introduces the \textit{SpeechFeedback} dataset.

\subsection{Task Definition}

The speech dialogue judgment task aims to compare a pair of speech responses based on criteria including truthfulness, honesty, helpfulness, instruction following, and speech instruction following, guided by a textual instruction. 
Formally, given a judge model $J$, the pairwise evaluation process can be formulated as:
\begin{equation}
ER = J(Q, R_1, R_2)
\label{eq:eval}
\end{equation}
where 
$Q$ is the textual query or instruction, $R_i$ represents the $i_{th}$ speech response, and $ER$ is the evaluation result.
The evaluation result includes five independent comparison labels $C_{a_j}$ and textual explanations $E_{a_j}$, where aspect $a_j \in$ \{truthfulness, honesty, helpfulness, instruction following, speech instruction following\} and $C_{a_j} \in$ \{win, lose, tie\} represent $R_1$ is better, $R_2$ is better, or both responses are of comparable quality respectively. 

In our task definition, we utilize textual queries as input instead of spoken audio. This design choice is paramount for ensuring a fair and standardized evaluation across diverse S2S LLMs. 
Our rationale is twofold. 
Firstly, existing speech dialogue datasets typically incorporate both textual and spoken queries, thereby facilitating the direct use of text queries without incurring additional overhead.
Second, S2S LLMs are trained on synthetic speech generated by heterogeneous TTS models (e.g., SLAM-Omni uses CosyVoice1, while Kimi-Audio uses Kimi-TTS). To mitigate potential distributional discrepancies between training and test audio, we adopt text-based instructions to ensure a fair evaluation across different S2S models.

\subsection{Data Construction and Pre-processing}

\textit{SpeechFeedback} dataset comprises two components: semantic evaluation and acoustic instruction following evaluation. 

\subsubsection{Semantic Evaluation Data Construction}

The semantic dimension speech preference data are synthesized from UltraFeedback \cite{cui2023ultrafeedback} using 7 TTS models. 
UltraFeedback is a large-scale, fine-grained, and diverse preference dataset comprising 64k instructions, each with four LLM-generated responses rated by GPT-4 across four dimensions: instruction-following, truthfulness, honesty, and helpfulness, accompanied by scores and rationales.
We utilize these textual preference annotations and synthesize corresponding speech responses. 

To prepare the corpus for voice-centric interaction and evaluation, we implement a multi-stage filtering pipeline. First, mathematical expressions, code segments, and multilingual prompts are removed, and special characters are sanitized to preserve natural prosody in downstream synthesis. As shown in Fig. \ref{fig:data construction}a, the filtered responses are synthesized by six open-source TTS models and one commercial TTS model—CosyVoice \citep{du2024cosyvoice}, CosyVoice2 \citep{du2024cosyvoice2}, SparkTTS \citep{wang2025spark}, ChatTTS\footnote{\url{https://github.com/2noise/ChatTTS}}, F5-TTS \citep{chen-etal-2024-f5tts}, Index-TTS \citep{deng2025indextts}, and gpt-4o-mini-tts\footnote{\url{https://platform.openai.com/docs/models/gpt-4o-mini-tts}}.
Utterances exhibiting a high Whisper ASR word-error rate or a duration shorter than 0.2 s are discarded. Absolute quality scores are subsequently converted to pairwise win–loss labels, and their rationales are rewritten in comparative form via Qwen2.5-32B-Instruct. The combined text–speech sequence is capped at 4,096 tokens. Further preprocessing details appear in the supplementary material.

\subsubsection{Acoustic Evaluation Data Construction}

Voice interaction allows users to explicitly or implicitly control the vocal characteristics of an SLM's response. For instance, a user might explicitly request a faster speaking rate, a response in a different language, or a specific emotional tone. 
In implicit scenarios, users expect SLMs to exhibit empathy, highlighting the importance of acoustic dimensions in evaluating spoken dialogue.

This paper focuses on three primary categories of speaking style control: emotion, gender, and cartoon voices, as illustrated in Fig. \ref{fig:data construction}b. Furthermore, as shown in Fig. \ref{fig:data construction}c,  we structure acoustic evaluation around three task formats: explicit TTS, explicit dialogue, and implicit dialogue.
Each data instance comprises an instruction, the corresponding speech responses, and an acoustic evaluation label with its accompanying rationale.

\noindent \textit{i. Explicit TTS Task} 

\begin{itemize}
    \item \textit{Instruction:} The Explicit TTS task utilizes the Ultrafeedback dataset, from which we sample 1,000 instructions for each speaking style category. For each instruction, the text response to be synthesized is randomly selected from four candidate responses. This text is then combined with a prompt template, which is randomly chosen from a set of 20 templates per category generated by GPT-4o.
    \item \textit{Speech Responses:} We first define \textit{incorrect label set} for each control category. For emotion control, labels are categorized as positive (happy, surprised) or negative (sad, fearful, angry). The incorrect label set for a target emotion comprises emotions from the opposing category and a neutral emotion.
    For gender control, the incorrect label set consists solely of the other gender label.
    For cartoon voice control, the incorrect label set includes a neutral voice and all cartoon voices except the target voice.
    Each sample includes two synthesized speech responses: one with the correct label and one with an incorrect label randomly sampled from the \textit{incorrect label set}, with sampling probabilities of 8:1:1 for correct-correct, correct-incorrect, and incorrect-incorrect pairs, respectively.
    \item \textit{Acoustic Evaluation Label and Rationale:} The final evaluation label is assigned as win, lose, or tie based on the ground truth labels used for synthesis. The instruction, the two text responses, and the assigned label are then provided as input to Qwen2.5-32B to generate a corresponding rationale.
\end{itemize}

\noindent \textit{ii. Explicit Dialogue Task} 

We also sample 1,000 instructions for each speaking style category from the Ultrafeedback dataset. The instruction is formed by combining a user query with a randomly sampled speaking style template. Subsequently, two text responses are randomly selected. The remaining procedure for selecting control labels, synthesizing audio, and constructing the evaluation output is identical to that of the Explicit TTS task.
Additionally, 180 \textit{mixed} samples are created to jointly control emotion and gender, with instructions like: Respond to \texttt{<query>} in a happy female voice.

\noindent \textit{iii. Implicit Dialogue Task} 

For the implicit dialogue task, both the instructions and the text responses are generated by \texttt{GPT-4o}.
We employ the self-instruct framework \cite{wang-etal-2023-self-instruct} to curate and translate implicit emotion data from Kimi-GenTest \citep{ding2025kimi}, and use them as seed prompts to generate 500 evaluation samples.
The subsequent steps—control label selection, speech synthesis, and label construction—follow the same pipeline as the explicit TTS task. However, the rationales are uniquely structured: we prompt \texttt{GPT-4o} to first explain the implied emotional intent of the query and then provide a template-based description of the emotional tones of each candidate response.

In summary, the overall acoustic evaluation data construction framework incorporates both semantic and acoustic preference annotation. The semantic labels and their corresponding rationales are annotated following the methodology detailed in the preceding section.

\section{Preliminary}

\subsection{Training Objective}
We explore two prominent fine-tuning approaches: Instruction Tuning (IT) and Reinforcement Learning (RL).
IT is a primary method for model fine-tuning, widely employed to learn specific output formats and align with human preferences.
On-policy RL methods have recently been shown to be particularly effective for preference learning from a limited number of samples. 
Moreover, RL tends to perform well where verification is simple, yet generation is complex \citep{li2025reinforcement, swamy2025all}. 
Evaluation task, which requires assigning a single discrete label (win, lose, or tie), presents a significant generation-verification gap. This makes it an ideal setting to investigate the effectiveness of RL in the speech modality.
The specific training objectives for each method are detailed below.
\paragraph{Instruction Tuning:} During the supervised fine-tuning (SFT) stage, the model is trained to minimize the discrepancy between generated responses and reference responses. Specifically, the training objective is to maximize the log-likelihood of the reference output sequence:
\begin{equation}
\mathcal{L}_{\text{SFT}} = - \mathbb{E}_{(x, y) \sim \mathcal{D}} \left[ \sum_{t=1}^{T} \log P_\theta(y_t \mid y_{<t}, x) \right]
\end{equation}
We explore two SFT settings based on the composition of the reference response $y$. In the SFT (label-only) setting $y$ consists solely of the correct judgment label, while in the SFT (with-rationale) setting $y$ comprises both the label and a corresponding rationale previously generated by GPT-4.

\paragraph{Reinforcement Learning:} We leverage the Group Relative Policy Optimization (GRPO) algorithm \citep{shao2024deepseekmath}, which maximizes a reward-weighted objective using a clipped policy ratio and a KL-divergence penalty to a reference policy. 
We define a rule-based reward function as a supervision signal:
\begin{equation}
    r_i = \alpha \cdot R_{a}(\hat{s}, s) + \gamma \cdot R_f
\end{equation}
\noindent where $R_a$ is the accuracy reward that measures the discrepancy between the predicted score $\hat{s}$ and the ground truth score $s$ ($1 \leq \hat{s}, s \leq 5$), as defined in Eq. (\ref{eq:ra}). The format reward $R_f \in \{0, 1\}$ evaluates whether the model correctly formats its output using the \texttt{<think></think>} and \texttt{<answer></answer>} tags. Specifically, the weighted averaging parameters are set to $\alpha = 1.0$ and $\gamma = 0.5$. 
The advantage estimation and training objective follow the standard GRPO formulation.
\begin{equation}
    R_{a}(\hat{s}, s) = \exp\left( - \frac{(\hat{s} - s)^2}{2\sigma^2} \right), \quad 0 \le \left | \hat{s} - s \right | \le 4 
\label{eq:ra}
\end{equation}

\begin{table*}[htbp]
    \centering
    \small
 
    \begin{tabular}{c|cccc|cccc}
      \toprule
      \multirow{2}{*}{\textbf{Model}} & \multicolumn{4}{c|}{\textbf{Accuracy}} & \multicolumn{4}{c}{\textbf{Agreement}} \\
      \cline{2-9}
      \specialrule{0em}{1pt}{1pt}
      & \textbf{Hel.} & \textbf{Hon.} & \textbf{IF.} & \textbf{Tru.} & \textbf{Hel.} & \textbf{Hon.} & \textbf{IF.} & \textbf{Tru.} \\
      \midrule
      \multicolumn{9}{l}{\cellcolor{black!10} \textbf{\textit{I. Baseline of cascaded ASR and text language models}}} \\
      Whisper + GPT-4o & 62.09 & 55.86 & 65.75 & 61.54 & 75.27 & 74.27 & 77.02 & 74.91 \\
      Whisper + Qwen2.5-32B & 53.48 & 40.66 & 59.89 & 40.29 & 67.40 & 65.20 & 70.70 & 58.79 \\
      Whisper + PandaLM-7B & 54.03 & 34.43 & 53.85 & 41.03 & 67.58 & 58.61 & 64.01 & 59.16 \\
      Whisper + Qwen2.5-omni-7B & 35.90 & 25.82 & 38.28 & 31.14 & 48.35 & 48.72 & 47.80 & 49.36 \\
      Whisper + Qwen2.5-omni-3B & 25.09 & 14.65 & 29.86 & 18.87 & 34.25 & 31.87 & 39.56 & 31.69 \\
      \midrule
      \multicolumn{9}{l}{\cellcolor{black!10} \textbf{\textit{II. Baseline of Speech-to-text large language models direct inference with prompt}}} \\
      Qwen2-Audio-Base & 7.15 & 4.95 & 7.51 & 7.32 & 9.52 & 9.43 & 8.89 & 10.53 \\
      Qwen2-Audio-Instruct & 23.81 & 15.38 & 24.73 & 18.13 & 34.98 & 33.43 & 33.15 & 32.88\\
      Qwen2.5-omni-3B & 37.73 & 30.04 & 41.02 & 37.18 & 53.48 & 54.67 & 54.12 & 55.86\\
      Qwen2.5-omni-7B & 41.76 & 29.49 & 44.69 & 38.83 & 56.50 & 55.31 & 56.50 & 58.06 \\
      \midrule
      \multicolumn{9}{l}{\cellcolor{black!10} \textbf{\textit{III. Different finetuning versions of} \texttt{SageLM}}} \\
      Qwen2-Audio-Instruct-7B-SFT & 70.33 & 68.50 & 65.57 & 65.02 & 82.97 & 81.32 & 77.56 & 78.66 \\
      Qwen2.5-omni-3B-SFT & 70.33 & \textbf{73.08} & 67.40 & 68.68 & 82.60 & \textbf{85.26} & 78.57 & 81.05 \\
      Qwen2.5-omni-7B-SFT (SageLM) & \textbf{72.35} & 67.40 & \textbf{72.35} & \textbf{73.26} & \textbf{83.61} & 81.78 & \textbf{81.68} & \textbf{84.07}\\
      \bottomrule
    \end{tabular}
    \caption{Semantic evaluation of \texttt{SageLM} and competing baselines on a human-annotated test set. We report accuracy and agreement (\%) across four dimensions: Helpfulness (Hel.), Honesty (Hon.), Instruction Following (IF.), and Truthfulness (Tru.).}
    \label{tab:semantic_result}
\end{table*}

\definecolor{color2}{RGB}{098,190,166} 
\definecolor{color3}{RGB}{253,186,107} 
\definecolor{color4}{RGB}{235,096,070} 
\definecolor{color5}{RGB}{111,146,225} 
\definecolor{color6}{RGB}{175,135,220} 
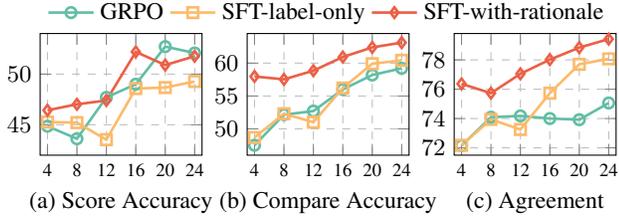
\begin{figure}[t]
\begin{tikzpicture}
\footnotesize{

\begin{axis}[
    name=plot1,              
    width=0.21\textwidth,
    height=0.18\textwidth,
    grid=major,
    grid style=dashed,
    xmin=3, xmax=25,
    ymin=42, ymax=54,
    xtick={4,8,12,16,20,24},
    xticklabels={\scriptsize{4}, \scriptsize{8},\scriptsize{12},\scriptsize{16},\scriptsize{20},\scriptsize{24}},
    xlabel={\footnotesize (a) Score Accuracy},
    xlabel style={yshift=0.5em},
    scaled ticks=false
]
    \addplot[color2, mark=o, line width=1pt] file {figs/fig_result/score/RL.txt}; 
    \addplot[color3, mark=square, line width=1pt] file {figs/fig_result/score/SFT-label.txt}; 
    \addplot[color4, mark=diamond, line width=1pt] file {figs/fig_result/score/SFT.txt}; 
\end{axis}

\begin{axis}[
    name=plot2,
    at={(plot1.north east)},   
    anchor=north west,         
    xshift=0.6cm,             
    width=0.21\textwidth,
    height=0.18\textwidth,
    grid=major,
    grid style=dashed,
    xmin=3, xmax=25,
    ymin=46, ymax=64.5,
    xtick={4,8,12,16,20,24},
    xticklabels={\scriptsize{4}, \scriptsize{8},\scriptsize{12},\scriptsize{16},\scriptsize{20},\scriptsize{24}},
    xlabel={\footnotesize (b) Compare Accuracy},
    xlabel style={yshift=0.5em},
    scaled ticks=false,
    legend columns=5,
    legend style={
      at={(0.50,1.0)}, 
      anchor=south,
      fill=none,
      draw=none,
      font=\footnotesize
    },
    legend entries={
      GRPO,
      SFT-label-only,
      SFT-with-rationale 
    }
]
    \addplot[color2, mark=o, line width=1pt] file {figs/fig_result/accuracy/RL.txt}; 
    \addplot[color3, mark=square, line width=1pt] file {figs/fig_result/accuracy/SFT-label.txt}; 
    \addplot[color4, mark=diamond, line width=1pt] file {figs/fig_result/accuracy/SFT.txt}; 
\end{axis}

\begin{axis}[
    name=plot3,
    at={(plot2.north east)},
    anchor=north west,
    xshift=0.6cm,
    width=0.21\textwidth,
    height=0.18\textwidth,
    grid=major,
    grid style=dashed,
    xmin=3, xmax=25,
    ymin=71.5, ymax=79.8,
    xtick={4,8,12,16,20,24},
    xticklabels={\scriptsize{4}, \scriptsize{8},\scriptsize{12},\scriptsize{16},\scriptsize{20},\scriptsize{24}},
    xlabel={\footnotesize (c) Agreement},
    xlabel style={yshift=0.5em},
    scaled ticks=false
]
    \addplot[color2, mark=o, line width=1pt] file {figs/fig_result/agreement/RL.txt}; 
    \addplot[color3, mark=square, line width=1pt] file {figs/fig_result/agreement/SFT-label.txt}; 
    \addplot[color4, mark=diamond, line width=1pt] file {figs/fig_result/agreement/SFT.txt};
\end{axis}

}
\end{tikzpicture}
    \caption{
    \label{fig:rl-sft} Preliminary: Reinforcement Learning versus Supervised Fine-Tuning on three evaluation metrics while training data scaling up (4k$\sim$24k $\times$ 4 aspects).}
\end{figure}

\subsection{RL or SFT? Empirical Analysis on Judgment Task}
DeepSeek R1 demonstrates that the reasoning capabilities of LLMs can be effectively enhanced via pure GRPO \citep{guo2025deepseek}. 
Recent studies further suggest that RL performs well in tasks where verification is easy but generation is challenging, such as open-domain QA \citep{swamy2025all} and audio multi-choice QA \citep{li2025reinforcement}. 
The task of evaluating speech responses exemplifies this generation-verification gap: verifying a judgment (win/lose/tie) is straightforward, whereas generating a high-quality judgment is non-trivial. This motivates our investigation into the comparative performance of GRPO and full-parameter SFT on speech response judgment task. In our preliminary experiments, both RL and SFT were trained for one epoch, with all other settings consistent with main experiments.

As shown in Fig. \ref{fig:rl-sft}(a), GRPO outperforms SFT (label-only) in scoring accuracy. WaveReward \citep{ji2025wavreward} also find this conclusion, showing that RL-based post-training surpasses LoRA-finetuned Qwen2.5-Omni. 
However, scoring accuracy is too strict to evaluation task, as even humans struggle to assign precise scores. In contrast, pair-wised comparison both better reflects human preferences and is easier to annotate.
Fig. \ref{fig:rl-sft}(b) and (c) indicate that GRPO and SFT perform comparably on compare accuracy, with SFT slightly outperforming GRPO on compare agreement.

We hypothesize that this difference stems from their learning objectives. GRPO's format reward encourages a reasoning process, whereas SFT (label-only) optimizes solely for the final label. To bridge this gap, we introduce an SFT-with-rationale setting, augmenting labels with explanations generated by GPT-4. As shown in Figure 3 (SFT-with-rationale), the inclusion of rationales improves the model’s understanding of evaluation results. Additionally, SFT with rationale outperform all other methods.

Furthermore, we observe that GRPO's lacks of supervision over the reasoning process leads to \textit{inconsistency between rationale and final result}. 
In a manual analysis of 100 sampled cases, the explanation contradicted the final score in 39\% of instances. This issue severely limits the reliability and explainability of the model as a speech judge.

\subsection{SageLM: Two-stage Training} 

To address the scarcity of acoustic preference data, we introduce a two-stage training strategy for \texttt{SageLM}. Drawing inspiration from curriculum learning \citep{bengio2009curriculum, wang2025rovrm}, this approach progressively builds the model's evaluative capabilities.

\textit{Semantic Preference Learning.} The first stage utilizes the abundant semantic preference annotations in the \textit{SpeechFeedback} dataset. This stage trains \texttt{SageLM} to evaluate responses along four core dimensions—truthfulness, honesty, helpfulness, and instruction following. Concurrently, the model learns to generate structured comparison outputs that include detailed explanatory rationales.

\textit{Acoustic Preference Learning.} Building upon the semantic foundation, the second stage incorporates a limited amount of acoustic preference data. This introduces a fifth evaluation dimension: speech instruction following. This dimension assesses whether the synthesized speech adheres to acoustic attributes. These attributes, such as emotion, gender, or a specific voice style, may be explicitly requested or implicitly implied by the user's query.

Informed by our preliminary experiments, both training stages adopt a rationale-augmented SFT method to supervise both the judgment labels and their corresponding explanations during post-training.

\section{Experiments}

\subsection{Experimental Setups}

\paragraph{Dataset} We employ the SpeechFeedback dataset, a synthetic dialogue evaluation corpus detailed previously. The training set consists of 316,544 semantic and 4,270 acoustic feedback instances. The test set contains 728 semantic and 410 acoustic evaluation instances, all manually verified to ensure label order aligns with human preferences.

\paragraph{Training and Generations Details} SageLM is trained using 8 NVIDIA A100-SXM4-80GB GPUs. For instruction tuning, we adopt full-parameter supervised fine-tuning with the following hyperparameters: 3 epochs, cutoff length of 4096, batch size of 16, learning rate of 1e-5, cosine learning rate scheduler with a warmup ratio of 0.1, and bf16 precision. 
Reinforcement learning is implemented using GRPO, which incurs higher training costs. Accordingly, we set: 1 epoch, group size of 8, cutoff length of 8192, batch size of 32 (4 instructions × 8 samples), learning rate of 1e-5, cosine scheduler with a 0.05 warmup ratio, and bf16 precision. Sampling during RL training uses: temperature = 1.0, top-p = 0.99, top-k = 50, and max completion length = 2048.
During inference, all models use identical decoding parameters: temperature = 0.95, top-p = 0.7, top-k = 50, and repetition penalty = 1.0. 
We report results averaged across three runs with seeds \{42, 123, 1234\} for reproducibility.

\begin{table*}[htbp]
    \centering
    \small
 
    \begin{tabular}{c|ccccc|ccccc}
      \toprule
      \multirow{2}{*}{\textbf{Model}} & \multicolumn{5}{c|}{\textbf{Accuracy}} & \multicolumn{5}{c}{\textbf{Agreement}} \\
      \cline{2-11}
      \specialrule{0em}{1pt}{1pt}
      & \textbf{Emo.} & \textbf{Gen.} & \textbf{Voi.} & \textbf{Imp.} & \textbf{Mixed} & \textbf{Emo.} & \textbf{Gen.} & \textbf{Voi.} & \textbf{Imp.} & \textbf{Mixed} \\
      \midrule
      \multicolumn{11}{l}{\cellcolor{black!10} \textbf{\textit{I. Baseline of Speech-to-text large language models direct inference with prompt}}} \\
      Qwen2-Audio-Base & 8.05 & 9.28 & 7.07 & 8.11 & 6.59 & 9.96 & 11.86 & 9.26 & 12.16 & 11.32 \\
      Qwen2-Audio-Instruct & 40.61 & 41.93 & 30.97 & 40.54 & 29.22 & 48.28 & 49.14 & 43.26 & 55.41 & 52.47 \\
      Qwen2.5-omni-3B & 44.45 & 41.24 & 40.74 & 37.84 & 24.69 & 53.64 & 49.14 & 53.87 & 52.70 & 48.97 \\ 
      Qwen2.5-omni-7B & 39.46 & 40.89 & 43.43 & 35.14 & 28.40 & 51.53 & 49.31 & 56.57 & 49.55 & 51.64 \\
      \midrule
      \multicolumn{11}{l}{\cellcolor{black!10} \textbf{\textit{II. Different finetuning versions of} \texttt{SageLM}}} \\
      Qwen2-Audio-Instruct-7B-SFT & 82.76 & 94.50 & 89.23 & 80.18 & 72.01 & 91.38 & 97.25 & 94.61 & 90.09 & 85.39 \\ 
      Qwen2.5-omni-3B-SFT & 91.18 & 96.22 & 94.95 & 94.59 & \textbf{90.12} & 95.59 & 98.11 & 97.47 & 97.47 & \textbf{94.44} \\ 
      Qwen2.5-omni-7B-SFT (SageLM) & \textbf{95.78} & \textbf{98.63} & \textbf{99.33} & \textbf{97.30} & 82.72 & \textbf{97.89} & \textbf{99.31} & \textbf{99.66} & \textbf{98.65} & 91.36 \\
      \bottomrule
    \end{tabular}
    \caption{Acoustic evaluation results against baselines on a human-annotated test set. We report accuracy and agreement (\%) across five dimensions: emotion (Emo.), gender (Gen.), voice (Voi.), implicit emotion (Imp.), and Mixed aspects (Mixed.) which jointly evaluate emotion and gender. Results for ASR baselines are not applicable as they ignore acoustic features.}
    \label{tab:acoustic_result}
\end{table*}

\paragraph{Evaluation Metrics} We report Accuracy and Agreement as primary metrics for both semantic and acoustic dimension evaluation. Accuracy follows Eq. (\ref{eq:eval}), measuring whether the model’s predicted comparison label matches the ground truth (win/lose/tie). Following PandaLM \citep{pandalm2024}, Agreement assigning 1 for complete agreement with human judgment, 0 for complete disagreement, and 0.5 otherwise. 
Final evaluation results are averaged over three inference runs.

\subsection{Experimental Results}

\paragraph{Comparison with Cascaded Baselines} The cascaded ASR-to-text evaluation pipeline is widely adopted in recent studies on evaluating S2S LLMs. Following prior work, we first transcribe speech using \texttt{whisper-large-v3-turbo} \citep{radford2022whisper}, then evaluate the transcribed text using several advanced text-based models: gpt-4o, the evaluation-specialized PandaLM-7B, and the Qwen family including Qwen2.5-32B and Qwen-omni-3B/7B.
As shown in Table \ref{tab:semantic_result} and \ref{tab:acoustic_result}, experimental results demonstrate that \textit{cascade pipelines serve as strong baselines for semantic-level evaluation}. 
Specifically, Whisper + gpt-4o pipeline achieves an average accuracy of 61.31\% and an average agreement rate of 75.37\% across four evaluation aspects. 
However, this approach lacks the capacity to evaluate acoustic dimensions, limiting its applicability.
In contrast, Whisper + Qwen2.5-Omni-3B/7B performs poorly due to lower parameter size and limited instruction-following capabilities. 
While few-shot prompting can typically guide output formatting, in QA settings with aroud 60s audio responses, context length becomes a bottleneck, limiting the usability of in-context learning (ICL).

\paragraph{Comparison with S2T LLM baselines} Inspired by evaluation practices in multi-modality LLMs, a common baseline for assessing generative models is to leverage unimodal understanding models, e.g. vision-LLMs for evaluating text-to-image generation. In the speech domain, S2T LLMs possess comprehensive speech understanding capabilities, making them natural candidates for evaluating spoken dialogue. Thus, we examine the evaluation capability of S2T LLMs, including Qwen-Audio-7B-base\&instruct and Qwen2.5-omni-3B/7B, using prompt engineering to elicit dialogue-level judgments.
Qwen2-Audio-Base, lacking instruction tuning, fails to follow task instructions in 82\% of cases, often repeating prompts or outputting [1\textbar 2\textbar Tie].
Although Qwen2.5-Omni outperforms Qwen2-Audio-Instruct-7B in evaluation tasks, it still falls short compared to the Whisper + gpt-4o cascade pipeline.
Context length limitation is more pronounced during inference with S2T LLMs: spoken input typically results in significantly longer token sequences than text, leaving no room for ICL.
In summary, current S2T models \textit{exhibit limited capability in both semantic and acoustic judgment}, with an average accuracy of only around 40\% in acoustic evaluation.

\paragraph{Different Finetuning Versions} 
The performance of the base model significantly influences the performance after post-trained. We compare the outcomes of two-stage training based on three models: Qwen2-Audio-Instruct-7B, Qwen2.5-omni-3B, and Qwen2.5-omni-7B. Experimental results demonstrate that the two-stage post-training enhances the speech dialogue evaluation ability across all models, with Qwen2.5-omni-7B achieving the best overall performance.
Ultimately, \textit{SageLM outperform both cascaded and SLM-based baselines} by at least 7.42\% and 26.20\% respectively, reaching 82.79\% agreement with human judgments.
Additional human evaluation confirms that the rationales generated by \texttt{SageLM} offer improved explainability and align with final results at 90.89\% by human evaluation.

\subsection{Analysis}

\definecolor{ugreen}{RGB}{098,190,166}
\definecolor{uyellow}{RGB}{253,186,107}
\definecolor{ured}{RGB}{235,096,070}
\definecolor{upurple}{RGB}{175,135,220} 

\begin{figure}[t]
\centering
\pgfplotsset{
   width=0.5\textwidth,
   height=0.23\textheight,
   symbolic x coords={1,2,3,4},
   enlarge y limits={upper,value=0.05},
   legend style={
      fill,
      at={(0, 0)},
      legend columns=2,
      legend cell align=left,
      anchor=south
      },
   }
\footnotesize{
\begin{tikzpicture}

\begin{axis}[
    at={(0em,1em)},
    ymajorgrids,
    grid style=dashed,
    legend entries={Qwen2.5-omni-3B, $+$ stage1 (RL$\dagger$), $+$ stage1 (SFT), $+$ stage1 + stage2 (SFT)},
    legend cell align={left},
    ybar,
    enlarge x limits=0.18,
    xtick align=inside,
    bar width=1.0em,
    xmax=4,
    xmin=1,
    xtick=data,
    nodes near coords, 
    nodes near coords style={
        font=\bfseries\tiny, 
        /pgf/number format/.cd, 
            fixed,             
            precision=1,      
        /tikz/.cd,             
        yshift=0.1pt             
    },
    nodes near coords align={vertical},
    ymin=20,
    ymax=99,
    ytick={30,40,50,60,70,80,90,100},
    yticklabels={30,40,50,60,70,80,90,100},
    xticklabels={Semantic\\Accuracy, Semantic\\Agreement, Acoustic\\Accuracy, Acoustic\\Agreement},
    xticklabel style={align=center,text width=2.5cm},
    yticklabel pos=left,
    ylabel style={yshift=-3em},xlabel style={yshift=0.3em,align=center},
    yticklabel style={/pgf/number format/fixed,/pgf/number format/fixed},
    legend style={draw=none, line width=1pt, at={(0.5,1.05)}, anchor=south},
    xtick=data,
    axis on top=false,
    error bars/y dir=both,
    error bars/y explicit,
    error bars/error bar style={black, thin},  
  ]
\addplot[fill=ugreen!50,draw=ugreen!80, area legend] coordinates { 
    (1,36.49) +- (0,1.8)  
    (2,54.53) +- (0,1.1)  
    (3,37.79) +- (0,1.9)  
    (4,51.66) +- (0,0.8)  
};
\addplot[fill=uyellow!50, draw=uyellow!80, area legend] coordinates { 
    (1,59.21) +- (0,1.5) 
    (2,75.06) +- (0,1.1)  
    (3,28.25) +- (0,1.7) 
    (4,56.98) +- (0,1.2)  
};
\addplot[fill=ured!50, draw=ured!80, area legend] coordinates { 
    (1,70.61) +- (0,0.8)
    (2,81.99) +- (0,0.7) 
    (3,30.01) +- (0,2.2) 
    (4,57.93) +- (0,1.5) 
};
\addplot[fill=upurple!50, draw=upurple!80, area legend, every node near coord/.append style={xshift=3pt}] coordinates { 
    (1,69.87) +- (0,0.7)
    (2,81.87) +- (0,0.8)  
    (3,93.41) +- (0,0.6)  
    (4,96.62) +- (0,0.4)  
};

\end{axis}

\end{tikzpicture}
}
	\caption{Analysis of the impact of stage1 semantic evaluation training and stage2 acoustic evaluation training.}
 \label{fig:ablation}
\end{figure}
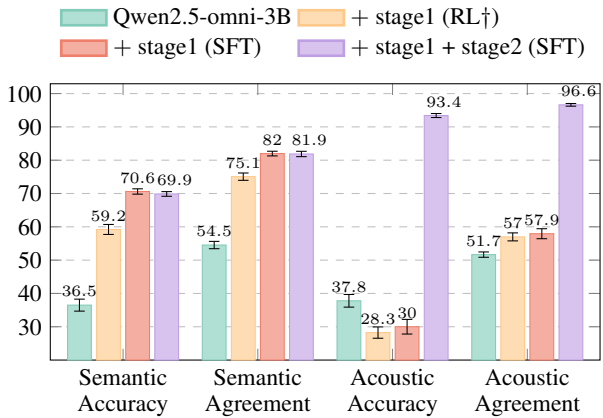

\paragraph{Two-stage Training}
Two-stage post-training substantially enhances the evaluation capability of S2T LLMs on speech responses. 
We conduct ablation studies to further understand its effects.
Fig. \ref{fig:ablation} 
shows that stage1, whether using GRPO or SFT, improves semantic evaluation but degrades acoustic evaluation. In contrast, stage2 incorporates both semantic and acoustic aspect preference data, thereby preserving semantic performance while significantly enhancing acoustic evaluation.
Each result is obtained by averaging over four semantic aspects or five acoustic aspects.

\paragraph{Position Consistency}
LLM-as-a-judge suffers from \textit{positional bias}, wherein the model exhibits a propensity to favor certain positions \cite{wang-etal-2024-large-language-models-fair, zheng2024judging}. Therefore, traditional evaluation requires assessing both orders of response pairs, effectively doubling inference cost.
Experiments demonstrate that \texttt{SageLM} offers more robust evaluations, exhibiting strong position consistency under response order reversal. On the \textit{SpeechFeedback} test dataset, \texttt{SageLM} achieves consistency rates of 84.95\% and 96.10\% on the semantic and acoustic dimensions respectively, confirming both its robustness and evaluation efficiency.

\definecolor{ugreen}{RGB}{098,190,166}
\definecolor{uyellow}{RGB}{253,186,107}
\definecolor{ured}{RGB}{235,096,070}
\definecolor{upurple}{RGB}{175,135,220} 
\begin{figure}[t]
\centering
\begin{tikzpicture}
\centering
\footnotesize{
\begin{axis}[
at={(0,0)},
    ymajorgrids,
    grid style=dashed,
    width=0.45\textwidth, height=0.22\textwidth,
    legend entries={Human, GPT-4o},
    legend style={
        fill opacity=0.5,
        text opacity=1,
        draw=none,
        line width=1pt,
        at={(axis cs:42,0.875)}, 
        anchor=south west,
        nodes={scale=0.8, transform shape},
        column sep=5pt,
        legend columns=2
    },
    xmin=8, xmax=105,
    ymin=0.72, ymax=0.93,
    xtick={10,20,...,100},
    xticklabels={10s,20s,30s,40s,50s,60s,70s,80s,90s,100s},
    ylabel=\footnotesize{Agreement},
    xlabel style={yshift=0.1em,align=center},
    ytick={0.75, 0.8, 0.85, 0.9},
    yticklabels={0.75, 0.8, 0.85, 0.9},
    scaled ticks=false,
    ]
    \addplot[ugreen, mark=square, line width=1pt] coordinates {(10,0.8690) (20,0.8967) (30,0.8095) (40,0.8029) (50,0.7946) (60,0.8182) 
    };
    \addplot[ured, mark=o, line width=1pt] coordinates {(10,0.91) (20,0.81) (30,0.88) (40,0.83) (50,0.83) (60,0.84) (70,0.79) (80,0.82) (90,0.78) (100,0.83)};
\end{axis}
}
\end{tikzpicture}
    \caption{Agreement vs. Combined responses pairs length.}
    \label{fig:ablation_length}

\end{figure}
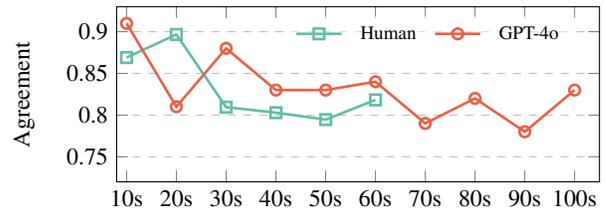

\paragraph{Model Performance of Difference Response Length}
As shown in Figure \ref{fig:ablation_length}, our analysis of the 728-sample test set indicates that SageLM's agreement with human annotations slightly decreases as the combined length of the evaluated response pair increases (pairs $>$70s are excluded due to scarcity). 
To corroborate this finding, we conducted a supplementary analysis on a larger dataset, sampling 100 examples from distinct intervals of combined response length. Experiment results show a similar slight decline in agreement between model's evaluations and GPT-4o annotations, though \textit{the agreement rate remained high at about 80\%}.

\paragraph{General vs. Task-specific Models in Evaluation Tasks}
Task-specific models for judgment is necessary in both text and speech response modality.
Firstly, Whisper + PandaLM-7B and Whisper + Qwen2.5-32B exhibit comparable performance, suggesting that \textit{task-specific evaluation models} can match the performance of general-purpose models with several times more parameters. 
Secondly, All baseline models showed significant improvements in their ability to evaluate speech responses after two-stage post-training, reaching a usable level with high agreement with human preference.

\paragraph{Out-of-Distribution Generalization on Real S2S LLMs Outputs}

\begin{table}[t]
    \centering
    \small
    \begin{tabular}{lcc}
    \toprule[0.7pt]
        \textbf{Model} & \textbf{Agreement} & \textbf{Accuracy} \\ \hline
        \specialrule{0em}{1pt}{1pt}
        Whisper + GPT-4o & 69.30\% & 53.80\% \\ 
        SageLM & 87.97\% & 81.01\% \\ 
    \toprule[0.7pt]
    \end{tabular}
    \caption{\label{OOD}
    Model evaluation agreement with human annotation on unseen AlpacaEval text-dataset.}
\end{table}

To assess the generalization of \texttt{SageLM} to \textit{unseen dataset} and \textit{real S2S LLMs output distribution}, we evaluate it on AlpacaEval followed by VoiceBench \cite{chen2024voicebench}, comparing Kimi-Audio with Qwen2.5-omni.
As shown in Table \ref{OOD}, \texttt{SageLM} surpasses the Whisper + GPT-4o pipeline by 18.67\% in agreement and 27.21\% in accuracy, validating its capability to evaluate within the true generative distribution of S2S LLMs.

To better understand this gap between \texttt{SageLM} and cascaded baseline, we further analyzed the evaluation results produced by Whisper in combination with GPT-4o. As illustrated in the case study in Appendix V\ref{sec:case}, Whisper introduced substantial transcription errors for Kimi-Audio responses due to speaking rate and prosody, which consequently biased GPT-4o toward favoring Qwen2.5-Omni. In contrast, \texttt{SageLM} and human evaluators were able to accurately capture the semantic content of the responses and make correct judgments.
This finding highlights the issue of error propagation in cascaded systems and underscores the necessity of an end-to-end model for speech judgement.

\begin{figure}[t]
\centering
\includegraphics[width=0.48\textwidth]{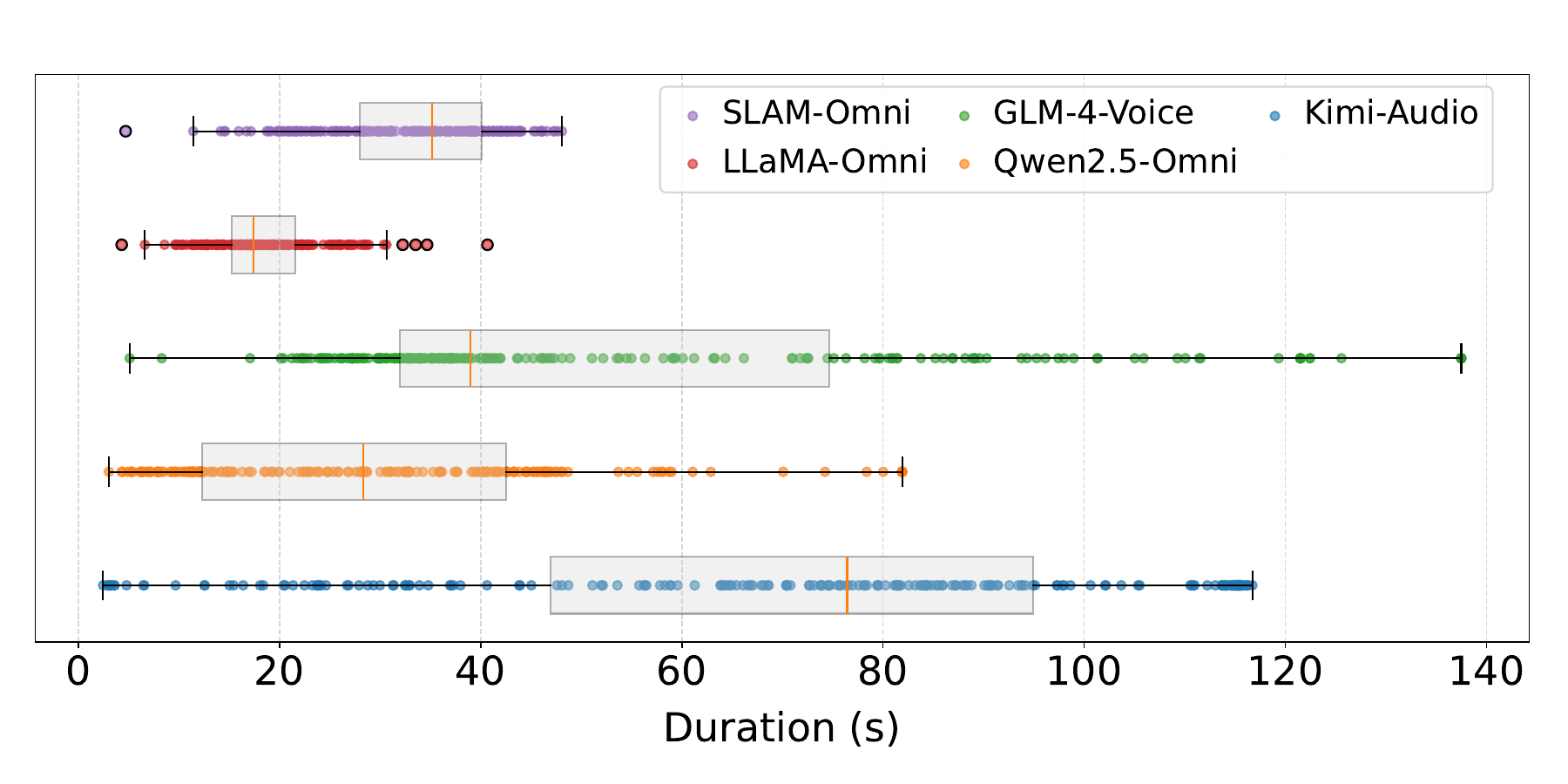} 
\caption{Response length distribution of S2S LLMs in AlpacaEval test dataset.
}
\label{fig:models_length}
\end{figure}

\definecolor{ugreen}{RGB}{098,190,166}
\definecolor{uyellow}{RGB}{253,186,107}
\definecolor{ured}{RGB}{235,096,070}
\definecolor{upurple}{RGB}{175,135,220} 
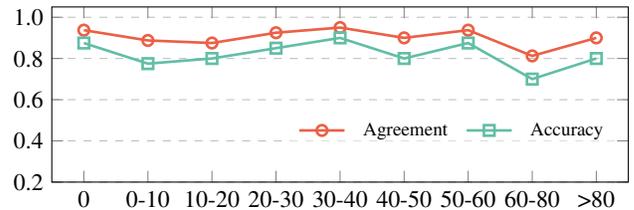
\begin{figure}[t]
\centering
\begin{tikzpicture}
\centering
\footnotesize{
\begin{axis}[
at={(0,0)},
    ymajorgrids,
    grid style=dashed,
    width=0.52\textwidth, height=0.22\textwidth,
    legend entries={Agreement, Accuracy},
    legend style={
        fill opacity=0.5,
        text opacity=1,
        draw=none,
        line width=1pt,
        at={(axis cs:32,0.36)},
        anchor=south west,
        nodes={scale=0.8, transform shape},
        column sep=5pt,
        legend columns=2
    },
    xmin=-5, xmax=85,
    ymin=0.2, ymax=1.05,
    xtick={0,10,20,...,80},
    xticklabels={0,0-10,10-20,20-30,30-40,40-50,50-60,60-80,>80},
    xlabel style={yshift=0.1em,align=center},
    ytick={0.2,0.4,0.6,0.8,1.0},
    yticklabels={0.2,0.4,0.6,0.8,1.0},
    scaled ticks=false,
    ]
    \addplot[ured, mark=o, line width=1pt] coordinates {(0,0.9375) (10,0.8875) (20,0.875) (30,0.925) (40,0.95) (50,0.9) (60,0.9375) (70,0.8125) (80,0.9)
    };
    \addplot[ugreen, mark=square, line width=1pt] coordinates {(0, 0.875) (10,0.775) (20,0.8) (30,0.85) (40,0.9) (50,0.8) (60,0.875) (70,0.7) (80,0.8)};
\end{axis}
}
\end{tikzpicture}
    \caption{Agreement vs. Combined responses pairs cutoff length. Horizontal axis in seconds.}
    \label{fig:cutoff_length}

\end{figure}

\paragraph{Model Performance of Difference Response Pair Cutoff Length}

SageLM’s 60s comparison window suffices for most evaluation settings.
On the out-of-domain AlpacaEval set, we measured speech response durations from five S2S LLMs as shown in Fig. \ref{fig:models_length}. Although SageLM truncates both compared audios to 60s, responses from SLAM-Omni, Llama-Omni and the upper quartile for Qwen2.5-Omni are all below 60 s. The 60s limit also covers the vast majority of longer generations from GLM-4-Voice and Kimi-Audio.

To further assess performance on long speech responses, we truncated GLM-4-Voice and Kimi-Audio outputs and analyzed the impact of truncated speech length on the agreement of \texttt{SageLM} and human preference. As shown in Fig. \ref{fig:cutoff_length}, agreement is stable when the total truncated speech length of the pair is $\le$ 60 s; but declines beyond this range, suggesting \texttt{SageLM} should be used more cautiously for longer inputs.

\section{Conclusions}
In this work, we investigate evaluation methods for spoken dialogue models and propose \texttt{SageLM}, motivated by two main challenges: (1) human-annotated evaluation is costly and difficult to scale, and (2) cascaded ASR-to-text-LLM pipelines suffer from error propagation and loss of acoustic information. To address these issues, we construct SpeechFeedback, a dataset comprising both semantic and acoustic evaluations of speech responses.
Preliminary experiments demonstrate that supervised fine-tuning of speech-to-text LLMs using interpretable feedback outperforms both label-only SFT and GRPO, even under a significant generation-verification mismatch in the evaluation task. To mitigate the scarcity of acoustic supervision, we adopt a two-stage SFT strategy to separately enhance semantic and acoustic evaluative capabilities.
Results show that \texttt{SageLM} achieves 82.79\% agreement with human judgments, surpassing cascaded and SLM-based baselines by at least 7.42\% and 26.20\%, respectively. 

\section{Future Work}
In this paper, we introduced \texttt{SageLM}, an end-to-end framework for spoken dialogue evaluation. Despite its promising results, our work has several limitations. Addressing these challenges presents valuable opportunities for advancing the spoken dialogue community. We highlight the following potential directions for future research:
\paragraph{Multi-turn Dialogue:} Extending \texttt{SageLM} to multi-turn conversations will require sophisticated mechanisms for managing and incorporating spoken dialogue history, which remains a significant design challenge.
\paragraph{Multilingual Evaluation:} Expanding \texttt{SageLM}'s capabilities to multilingual contexts is hindered by two primary factors: the scarcity of datasets and the limitations of current TTS technologies, particularly in realistically synthesizing code-switched utterances.
\paragraph{Full-duplex Interaction:} Evaluating full-duplex conversations \cite{arora2025talking, ge2025flexi} introduces substantial difficulties. Beyond the persistent issues of data scarcity and error propagation, this domain presents new challenges, such as dual-stream (user and model) speech understanding and multi-speaker scenarios.

\section{Acknowledgments}
This work was supported in part by the National Science Foundation of China (Nos. 62276056 and U24A20334), the Yunnan Fundamental Research Projects (No.202401BC070021), the Yunnan Science and Technology Major Project (No. 202502AD080014), and the Program of Introducing Talents of Discipline to Universities, Plan 111 (No.B16009).

\bibliography{aaai2026}

\section{Appendix}

\definecolor{PromptBoxTitleColor}{RGB}{76,76,76}
\definecolor{PromptBoxColor}{RGB}{247,247,255}
\definecolor{block-gray}{gray}{0.85}

\newtcolorbox{mybackground}{colback=block-gray,grow to right by=0mm,grow to left by=0mm,boxrule=0pt,boxsep=0pt,breakable}

\subsection{I. Semantic Preference Annotations}

\subsubsection{Dataset Filtering Pipeline}
~\\
To adapt this data for voice-centric interaction and evaluation, we apply a multi-stage filtering pipeline. 

First, we excluded instructions related to \textit{math and code} using Qwen2.5-32B-Instruct \cite{qwen2.5report}, as these domains typically involve highly structured content (e.g., formulas, SQL, HTML, LaTeX, Python, C++) that pose challenges for voice modality due to their reliance on precise formatting and extended context.

Second, since UltraFeedback includes \textit{multilingual} instructions (e.g., translation tasks), and current TTS models struggle with code-switching across languages, we applied a language detection tool\footnote{\url{https://github.com/Mimino666/langdetect}} to remove such instances.

Third, dialogue data often contains special characters (e.g., \texttt{\textbackslash n}) that disrupt natural sentence boundaries and prosodic phrasing in speech. We replaced these with appropriate punctuation to improve speech segmentation.
Then, six open-source TTS models and one commercial TTS model were used to synthesize the filtered semantic composition data, namely CosyVoice \citep{du2024cosyvoice}, CosyVoice2 \citep{du2024cosyvoice2}, SparkTTS \citep{wang2025spark}, ChatTTS\footnote{\url{https://github.com/2noise/ChatTTS}}, F5-TTS \citep{chen-etal-2024-f5tts}, Index-TTS \citep{deng2025indextts}, and gpt-4o-mini-tts\footnote{\url{https://platform.openai.com/docs/models/gpt-4o-mini-tts}}.

Furthermore, we filtered out samples with high \textit{Whisper ASR WER}, as these typically indicate degraded TTS quality or the presence of residual mathematical, programming, or multilingual content. 
Specifically, the threshold $\theta_{WER}$ is set to 0.2 when the word count $l$ of the transcribed text is less than 400 and linearly decreases to 0 as $l$ increases to 600 , which can be expressed as:
\begin{equation}
    \theta_{WER} = 
    \begin{cases}
        0.2, & \text{if } l < 400\\
        - \dfrac{l} {1000} + 0.6, & \text{if } 400 \leq l < 600\\
        0, & \text{if } l \geq 600
    \end{cases}
\end{equation}

Additionally, we discard any speech responses with a duration of less than 0.2 seconds.
Finally, we transform absolute ratings into relative win-loss labels. The rationales for these ratings are rewritten into a comparative format using Qwen2.5-32B-Instruct, with the specific prompt detailed in the next section.
We also ensure that the combined token \textit{length} of text and speech did not exceed 4096 tokens.\\\\

\subsubsection{Data Filtering Template}
~\\
Below is the prompt used to identify and exclude examples related to math and code.
\begin{mybackground}
You need to determine whether the instruction-response pair pertain to a mathematical or coding task. Do not respond to the instruction, just output Yes or No.\\
\\
\textasciigrave \textasciigrave \textasciigrave \\
Instruct: \\
Trying to create an alt character. Which prompt should I use?\\
Response: \\
The response is confidently incorrect, as it does not address the instruction and seems to misunderstand the question.\\
\textasciigrave \textasciigrave \textasciigrave \\
\textbackslash boxed\{No\}\\
\\
\textasciigrave \textasciigrave \textasciigrave \\
Instruct: \\
for f(x)=(3x-9)/(x\^{}2-7x+12), determine the vertical asymptote, horizontal asymptote, x intercept, and y intercept and explain your reasoning.\\
Response: \\
Vertical asymptote: There is no vertical asymptote in this rational function.\\
Horizontal asymptote: The equation of y=3/5 represents the horizontal asymptote. This can be determined by removing all terms that include x and then dividing the numerator and denominator by the highest power of x, which in this case is x.\\
x-intercepts: The x-intercepts occur when y=0, thus, we set f(x)=0 and solve for x. This results in x=3, which is the only real value of x that makes the denominator equal to zero.\\
y-intercept: The value of f(x) when x=0 is 12, which is the y-intercept.\\
Explanation: When examining the function f(x), we can see that as x approaches infinity, the value of f(x) approaches 3/5, which is its horizontal asymptote. Additionally, we can see that when x approaches 3 or 3, the value of f(x) approaches infinity, which means that there is no vertical asymptote. Finally, the x-intercepts occur when y=0, which results in x=3, and the y-intercept is obtained by evaluating f(0) (y-value at x=0), which is 12.\\
\textasciigrave \textasciigrave \textasciigrave \\
\textbackslash boxed\{Yes\}\\\\
\textasciigrave \textasciigrave \textasciigrave \\
Instruct:\\
\{instruct\}\\
Response:\\
\{response\}
\\
\textasciigrave \textasciigrave \textasciigrave \\
\end{mybackground}

\subsection{III. Acoustic Preference Annotations}
We first present an example that prompts GPT-4o to generate instruction templates for emotion control in speech conversation. Then we provide the instruction templates used in each acoustic instruction-following evaluation task.

\subsubsection{GPT-4o prompt}
~
\begin{mybackground}
Generate 20 diverse one-line English instruction templates for controlling speaking emotion in a dialogue system. 

The templates will be placed at the beginning of the dialogue session. 

Each template must include the placeholder <emotion> to specify the required emotion.

The templates should be declarative instructions, not questions. 

The emotions are: happy, sad, angry, surprise, fearful. 

Here is an example: Please answer the following question with a <emotion> tone.

Only output 20 templates, one per line, with no additional explanation. 

Ensure that the instruction flows naturally and smoothly when the <emotion> tag is replaced with an emotion.
\end{mybackground}

\subsubsection{Speech Conversation Voice Control}
~
\begin{mybackground}
Respond using \textless character\textgreater's voice.\\
Convey your message with \textless character\textgreater's voice.\\
Share your thoughts in \textless character\textgreater's voice.\\
Speak using \textless character\textgreater's voice.\\
Communicate in \textless character\textgreater's voice.\\
Deliver your response with \textless character\textgreater's voice.\\
Provide your answer in \textless character\textgreater's voice.\\
Express your ideas with \textless character\textgreater's voice.\\
Project your speech with \textless character\textgreater's voice.\\
Respond with \textless character\textgreater's voice.\\
Speak as if you were \textless character\textgreater.\\
Channel a \textless character\textgreater's voice in your words.\\
Let your words have a \textless character\textgreater's voice.\\
Adjust your speech to sound like \textless character\textgreater.\\
Craft your response with a \textless character\textgreater's voice.\\
Speak while embodying a \textless character\textgreater's voice.\\
Impart a \textless character\textgreater's voice in your speaking.\\
Infuse your dialogue with a \textless character\textgreater's voice.\\
Articulate yourself in a \textless character\textgreater's voice.\\
Speak like \textless character\textgreater.
\end{mybackground}

\subsubsection{TTS Voice Control}
~
\begin{mybackground}

Read this text using \textless character\textgreater 's voice.\\
Speak this text aloud using \textless character\textgreater 's voice.\\
Read this passage with \textless character\textgreater 's voice.\\
Use \textless character\textgreater 's voice to read this text.\\
Present this content with \textless character\textgreater 's voice.\\
Deliver this speech using \textless character\textgreater 's voice.\\
Render this material with \textless character\textgreater 's voice.\\
Produce the audio with \textless character\textgreater 's voice.\\
Read aloud this text with \textless character\textgreater 's voice.\\
Speak clearly with \textless character\textgreater 's voice.\\
Read the following text using \textless character\textgreater 's voice.\\
Speak the following text aloud with \textless character\textgreater 's voice.\\
Narrate the following passage with \textless character\textgreater 's voice.\\
Use \textless character\textgreater 's voice to narrate the text.\\
Present the following content with \textless character\textgreater 's voice.\\
Deliver the following speech using \textless character\textgreater 's voice.\\
Render the following material with \textless character\textgreater 's voice.\\
Produce audio for this text using \textless character\textgreater 's voice.\\
Read this passage clearly with \textless character\textgreater 's voice.\\
Speak this passage aloud using \textless character\textgreater 's voice.

\end{mybackground}

\subsubsection{Speech Conversation Emotion Control}
~
\begin{mybackground}
Respond using a \textless emotion\textgreater\ voice.\\
Convey your message with a \textless emotion\textgreater\ demeanor.\\
Share your thoughts in a \textless emotion\textgreater\ way.\\
Articulate with a \textless emotion\textgreater\ tone.\\
Communicate in a \textless emotion\textgreater\ manner.\\
Deliver your response with a \textless emotion\textgreater\ flavor.\\
Provide your answer in a \textless emotion\textgreater\ style.\\
Express your ideas with a \textless emotion\textgreater\ attitude.\\
Project your speech with a \textless emotion\textgreater\ sentiment.\\
Respond with a \textless emotion\textgreater\ inflection.\\
Speak as if you were feeling \textless emotion\textgreater.\\
Channel a \textless emotion\textgreater\ vibe in your words.\\
Let your words reflect a \textless emotion\textgreater\ state.\\
Adjust your speech to sound \textless emotion\textgreater.\\
Craft your response with a \textless emotion\textgreater\ essence.\\
Speak while embodying a \textless emotion\textgreater\ emotion.\\
Impart a \textless emotion\textgreater\ quality in your voice.\\
Infuse your dialogue with a \textless emotion\textgreater\ spirit.\\
Articulate yourself in a \textless emotion\textgreater\ fashion.\\
Let your expression resonate with a \textless emotion\textgreater\ mood.\\
\end{mybackground}

\subsubsection{TTS Emotion Control}
~
\begin{mybackground}
Read this text with a \textless emotion\textgreater\ voice.\\
Speak the following passage with a \textless emotion\textgreater\ tone.\\
Deliver this message using a \textless emotion\textgreater\ expression.\\
Recite the content with a \textless emotion\textgreater\ feeling.\\
Render the text in a \textless emotion\textgreater\ voice.\\
Perform the passage with a \textless emotion\textgreater\ tone.\\
Express the following using a \textless emotion\textgreater\ voice.\\
Read the following with a \textless emotion\textgreater\ inflection.\\
Speak this text aloud with a \textless emotion\textgreater\ feeling.\\
Deliver the message in a \textless emotion\textgreater\ voice.\\
Recite this passage with a \textless emotion\textgreater\ expression.\\
Render the following text with a \textless emotion\textgreater\ tone.\\
Perform the content using a \textless emotion\textgreater\ voice.\\
Express the message with a \textless emotion\textgreater\ feeling.\\
Read aloud the following text with a \textless emotion\textgreater\ tone.\\
Speak the content with a \textless emotion\textgreater\ voice.\\
Deliver this passage with a \textless emotion\textgreater\ inflection.\\
Recite the message using a \textless emotion\textgreater\ voice.\\
Render the content with a \textless emotion\textgreater\ feeling.\\
Perform the following text with a \textless emotion\textgreater\ voice.\\
\end{mybackground}

\subsubsection{Speech Conversation Gender Control}
~
\begin{mybackground}
Answer this question with a male voice.\\
Respond using a man's voice.\\
Provide your answer in a male tone.\\
Use a man's voice to reply to this question.\\
Give your response with a male voice.\\
Answer using a male narrator's voice.\\
Respond with a man's tone.\\
Provide the answer using a male speaker.\\
Reply with a confident male voice.\\
Answer naturally with a male voice.\\
Answer this question with a female voice.\\
Respond using a woman's voice.\\
Provide your answer in a soft female tone.\\
Use a feminine voice to reply to this question.\\
Give your response with a clear female voice.\\
Answer using a female narrator's voice.\\
Respond with a woman's tone.\\
Provide the answer using a female speaker.\\
Reply with a gentle female voice.\\
Answer naturally with a female voice.\\
\end{mybackground}

\subsubsection{TTS Gender Control}
~
\begin{mybackground}
Read this text with a male voice.\\
Speak the following using a man's voice.\\
Deliver this passage in a male tone.\\
Use a man's voice to present the text.\\
Repeat this content with a male voice.\\
Produce the speech using a male narrator.\\
Render this material with a man's tone.\\
Present the following using a male speaker.\\
Repeat this text with a male voice.\\
Speak naturally with a male voice for this passage.\\
Read this text with a female voice.\\
Speak the following using a woman's voice.\\
Deliver this passage in a soft female tone.\\
Use a feminine voice to present the text.\\
Narrate this content with a clear female voice.\\
Produce the speech using a female narrator.\\
Render this material with a woman's tone.\\
Present the following using a female speaker.\\
Repeat this text with a female voice.\\
Speak naturally with a female voice for this passage.\\
\end{mybackground}
\subsubsection{Speech Conversation Gender \& Emotion Control}
~
\begin{mybackground}
Answer this question with a male voice and a \textless emotion\textgreater~tone.\\
Respond using a man's voice with a \textless emotion\textgreater~inflection.\\
Provide your answer in a male voice with a \textless emotion\textgreater~tone.\\
Use a man's voice and a \textless emotion\textgreater~tone to reply.\\
Speak naturally with a male voice and a \textless emotion\textgreater~tone.\\
Deliver your response in a male voice with a \textless emotion\textgreater~inflection.\\
Answer with a man's voice and a \textless emotion\textgreater~tone.\\
Respond using a male voice with a \textless emotion\textgreater~tone.\\
Give your answer in a man's voice with a \textless emotion\textgreater~tone.\\
Provide the reply with a male voice and a \textless emotion\textgreater~tone.\\
Answer this question with a female voice and a \textless emotion\textgreater~tone.\\
Respond using a woman's voice with a \textless emotion\textgreater~inflection.\\
Provide your answer in a female voice with a \textless emotion\textgreater~tone.\\
Use a woman's voice and a \textless emotion\textgreater~tone to reply.\\
Speak naturally with a female voice and a \textless emotion\textgreater~tone.\\
Deliver your response in a female voice with a \textless emotion\textgreater~inflection.\\
Answer with a woman's voice and a \textless emotion\textgreater~tone.\\
Respond using a female voice with a \textless emotion\textgreater~tone.\\
Give your answer in a woman's voice with a \textless emotion\textgreater~tone.\\
Provide the reply with a female voice and a \textless emotion\textgreater~tone.\\
\end{mybackground}

\subsubsection{TTS Gender \& Emotion Control}
~
\begin{mybackground}
Read this text with a male voice and a \textless emotion\textgreater~tone.\\
Speak the following using a man's voice with a \textless emotion\textgreater~inflection.\\
Deliver this passage in a male voice with a \textless emotion\textgreater~tone.\\
Use a man's voice to read this text with a \textless emotion\textgreater~tone.\\
Narrate this content with a male voice and a \textless emotion\textgreater~inflection.\\
Produce the speech using a male voice with a \textless emotion\textgreater~tone.\\
Render this material with a man's voice and a \textless emotion\textgreater~tone.\\
Present the following using a male voice with a \textless emotion\textgreater~inflection.\\
Repeat this text with a male voice and a \textless emotion\textgreater~tone.\\
Speak naturally with a male voice and a \textless emotion\textgreater~tone.\\
Read this text with a female voice and a \textless emotion\textgreater~tone.\\
Speak the following using a woman's voice with a \textless emotion\textgreater~inflection.\\
Deliver this passage in a female voice with a \textless emotion\textgreater~tone.\\
Use a woman's voice to read this text with a \textless emotion\textgreater~tone.\\
Narrate this content with a female voice and a \textless emotion\textgreater~inflection.\\
Produce the speech using a female voice with a \textless emotion\textgreater~tone.\\
Render this material with a woman's voice and a \textless emotion\textgreater~tone.\\
Present the following using a female voice with a \textless emotion\textgreater~inflection.\\
Repeat this text with a female voice and a \textless emotion\textgreater~tone.\\
Speak naturally with a female voice and a \textless emotion\textgreater~tone.\\
\end{mybackground}

\subsection{IV. SageLM Training Prompt}
The following is the default prompt template used in SageLM training. The \texttt{\{Aspect\}} denotes one of the evaluation dimensions including helpfulness, instruction following, honesty, truthfulness, emotion instruction following, gender instruction following, and character instruction following. The \texttt{\{Instruction\}} represents the textual instruction, and each \texttt{<audio>} tag serves as a placeholder for corresponding audio response.
\begin{mybackground}

Below are two responses for a given task. The task is defined by the Instruction. Evaluate in terms of **\{Aspect\}** and indicate a better response using 1, 2 or Tie.\\
\#\#\# Instruction:\\
\{Instruction\}\\\\
\#\#\# Response 1:\\\textless audio\textgreater\\\\
\#\#\# Response 2:\\\textless audio\textgreater\\
\end{mybackground}

\subsection{V. Template for S2T Baselines Inference}

We adopt the template blow to compare S2T LLM baselines, including Qwen2-Audio-Base, Qwen2-Audio-Instruct, Qwen2.5-omni-3B and Qwen2.5-omni-7B.

\begin{mybackground}
Below are two responses for a given task. The task is defined by the Instruction. Evaluate in terms of **\{Task\}** and indicate a better response using 1, 2 or Tie.\\
\#\#\# Instruction:\\
\{Instruction\}\\\\
\#\#\# Response 1:\\\textless audio\textgreater\\\\
\#\#\# Response 2:\\\textless audio\textgreater\\\\
The following is a format demonstration only.\\
Two responses are audio files.\\
\#\#\# Instruction:\\
\textasteriskcentered{}instruction for task\textasteriskcentered{}\\
\#\#\# Response 1:\\
\textasteriskcentered{}audio\textasteriskcentered{}\\
\#\#\# Response 2:\\
\textasteriskcentered{}audio\textasteriskcentered{}\\
Answer: \textless Answer\textgreater[1\textbar 2\textbar Tie]\textless /Answer\textgreater\\
This is a format example. Any output formatting errors will result in rejection.
\end{mybackground}

\subsection{VI. Case Study of LLMs-as-a-judge}
\label{sec:case}
\begin{table}[h]
  \centering
  \small
  \setlength{\tabcolsep}{4pt}            
  \renewcommand{\arraystretch}{1.05}      
  \begin{tabularx}{\columnwidth}{@{} l X @{}}
    \toprule
    \textbf{Model} & \textbf{Transcription} \\
    \midrule
    Kimi-Audio &
    Stress can be a real buzzkill, but there are some cool ways to tackle it. Here are some effective stress-busting strategies. E. Deep breathing. Take slow, deep breaths in through your nose and out through your mouth. It's like a mini-meditation. R. Exercise. Get moving for many exercises. Get moving for many times. Focus on the present moment. Pay attention to your thoughts, feelings, and surroundings. E. It's like a mental reset for progressive muscle relaxation. Self-meditation. Progressive muscle relaxation. Self-meditation. Focus on the most unrun. Family member. Read. A humorous book. Or hang out with people who make you laugh. It's like a stress-free mental vacation. 7. Get outdoors. Spend time in nature. Take a walk. Or simply gaze out. A window. It's like a stress-free, self-care bake. Remember, everyone is unique, so experiment with different techniques to find what works best for you. \\ \hline 
    Qwen2.5-omni &
    Well, you could try exercise. Just going for a jog or doing some yoga can really help. It releases endorphins which make you feel bad. Another thing is to get enough sleep. When you're well-rested, you can handle stress better. Also, talking to someone, like a friend or family member, can be great. They can give you advice or just listen. And don't forget about hobbies. If you like painting or reading, do that when you can. If you want to know more about dealing with stress or have other questions, feel free to ask. \\
    \bottomrule
  \end{tabularx}
  \caption{Case study: model transcriptions on a stress-management prompt.}
  \label{tab:case}
\end{table}

The speech responses from Kimi-Audio were substantially mis-transcribed by Whisper due to the hoarse timbre, which consequently biased GPT-4o toward favoring the outputs of Qwen2.5-Omni.

More details and audios are available in the GitHub project: \url{https://github.com/IronBeliever/SageLM}.

\end{document}